\documentclass[10pt,twocolumn,letterpaper]{article}

\usepackage[pagenumbers]{cvpr}      % To produce the REVIEW version
\definecolor{cvprblue}{rgb}{0.21,0.49,0.74}
\usepackage[pagebackref,breaklinks,colorlinks,allcolors=cvprblue]{hyperref}

\usepackage{xspace}
\usepackage{graphicx}
\usepackage{amssymb}
\usepackage{enumitem}
\usepackage{xcolor}
\usepackage{wrapfig}
\usepackage{booktabs}
\usepackage{multirow}
\usepackage{algorithm}
\usepackage{algpseudocode}
\usepackage{amsmath}

\definecolor{myred}{HTML}{B85450}
\definecolor{myblue}{HTML}{6C8EBF}
\definecolor{myyellow}{HTML}{D6B656}
\definecolor{mygreen}{HTML}{00CC00}

\def\ie{\emph{i.e.}}
\def\eg{\emph{e.g.}}

\newcommand{\LINA}[0]{\mbox{\textsc{Lina}}\xspace}

\title{\LINA: Learning INterventions Adaptively for Physical Alignment and Generalization in Diffusion Models}

\author{
    Shu Yu$^{1,2,3}$ \quad Chaochao Lu$^{1}$\thanks{Corresponding author.} \\
    $^{1}$Shanghai AI Laboratory \quad
    $^{2}$Shanghai Innovation Institute \quad
    $^{3}$Fudan University \\
    {\tt\small \{yushu, luchaochao\}@pjlab.org.cn}
}

\begin{document}
\maketitle
\begin{abstract}
Diffusion models (DMs) have achieved remarkable success in image and video generation. However, they still struggle with (1) physical alignment and (2) out-of-distribution (OOD) instruction following. We argue that these issues stem from the models' failure to learn causal directions and to disentangle causal factors for novel recombination.
We introduce the Causal Scene Graph (CSG) and the Physical Alignment Probe (PAP) dataset to enable diagnostic interventions. This analysis yields three key insights. First, DMs struggle with multi-hop reasoning for elements not explicitly determined in the prompt. Second, the prompt embedding contains disentangled representations for texture and physics. Third, visual causal structure is disproportionately established during the initial, computationally limited denoising steps.
Based on these findings, we introduce \textbf{LINA (Learning INterventions Adaptively)}, a novel framework that learns to predict prompt-specific interventions, which employs (1) targeted guidance in the prompt and visual latent spaces, and (2) a reallocated, causality-aware denoising schedule. Our approach enforces both physical alignment and OOD instruction following in image and video DMs, achieving state-of-the-art performance on challenging causal generation tasks and the Winoground dataset. Our project page is at \texttt{\small \url{https://opencausalab.github.io/LINA}}.
\end{abstract}    
\section{Introduction}
Diffusion models (DMs) are proposed to achieve both flexibility and tractability in generative modeling \citep{sohl2015deep}, by learning to predict the injected noise in the data point at any timestep \citep{ho2020denoising, song2020score, song2020denoising}. These models first achieved significant success in image generation \citep{ho2020denoising, song2020denoising, rombach2022high, ramesh2022hierarchical, saharia2022photorealistic, esser2024scaling}, and have since been extended to video \citep{ho2022video, blattmann2023stable, wan2025wan}, audio \citep{kong2020diffwave, liu2023audioldm, chen2024f5}, and text \citep{li2022diffusion, nie2025large}.

\begin{figure}[t]
    \centering
    \includegraphics[width=\columnwidth]{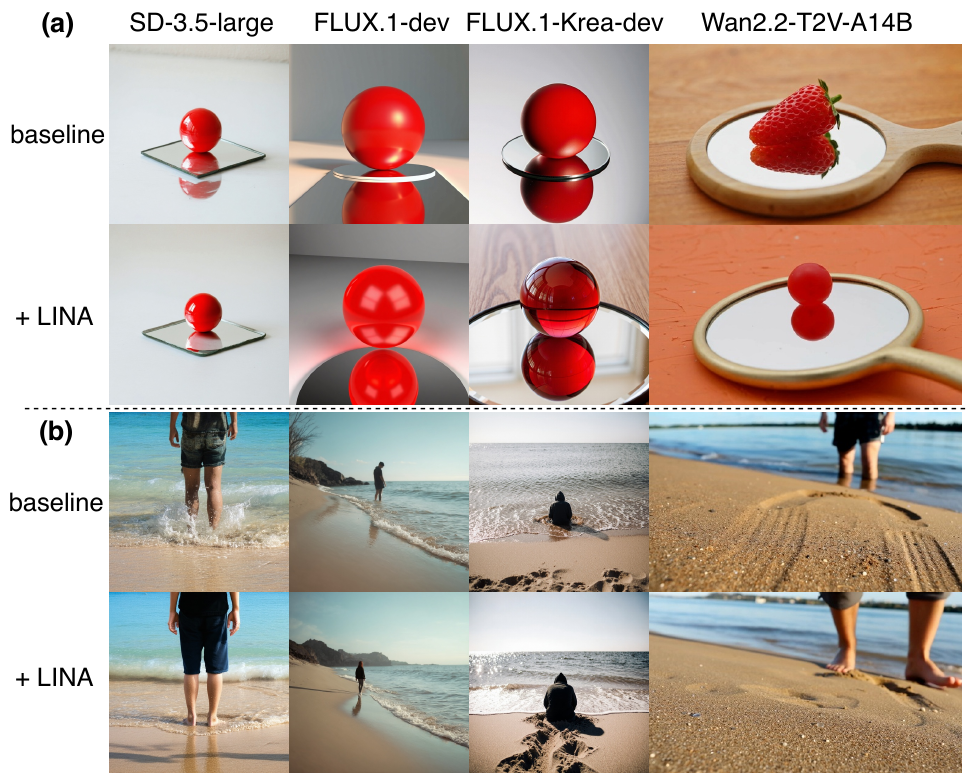}
    \caption{\textbf{Failures in DMs and \LINA's improvement.} (a) Generated with prompt: ``A red ball on a small mirror.'' Baseline models generate reflections extending beyond the mirror surface or produce texture errors. (b) Generated with Winoground \citep{thrush2022winoground} prompt: ``a person is \textit{close to} the water and \textit{in} the sand.'' Baseline models incorrectly place the person \textit{in} the water.}
    \label{fig:overall_demo}
\end{figure}

DMs for visual generation are widely considered a promising path toward realizing world models \citep{ding2025understanding}. However, despite their success, state-of-the-art (SOTA) DMs still struggle with two key challenges: (1) achieving \textbf{physical alignment} \citep{han2025can} and (2) following out-of-distribution (\textbf{OOD}) instructions \citep{ho2022classifier, chung2024cfg++, bradley2024classifier}.
We argue that these two issues stem from the DMs' deficient understanding of causality \citep{scholkopf2021toward,pearl2018book,han2025can,tong2025causal}, which manifests in two key failures: (1) modeling \textbf{symmetric correlations} rather than directional causality, preventing physical alignment, and (2) the \textbf{entanglement of causal factors}, hindering the novel recombination of visual elements for OOD generation.
As shown in Fig.~\ref{fig:overall_demo}, DMs misinterpret causal directions by producing redundant reflections in areas without mirrors (Fig.~\ref{fig:overall_demo}a), or entangle concepts such as ``person'' and ``water'', thus rendering the ``person in the water'' and failing to follow instructions (Fig.~\ref{fig:overall_demo}b).

To diagnose the root of these failures, we introduce a causal intervention-based analysis. First, we introduce the \textbf{Causal Scene Graph (CSG)}, a representation for modeling the generative process of DMs. Inspired by Causal Graphical Model (CGM) \citep{peters2017elements} and Scene Graph (SG) \citep{chang2021comprehensive}, CSG unifies causal dependencies and spatial layouts, providing a basis for our interventions. Second, based on CSG, we construct the \textbf{Physical Alignment Probe (PAP) dataset}, which consists of structured prompts, a corresponding set of images generated by SOTA DMs, and segmentation masks for elements in these images, designed to quantify DMs' physical alignment and OOD instruction following.

We conduct a diagnostic probing analysis on the PAP dataset using CSG-guided interventions, specifically masked inpainting conditioned on partial causal elements in the given image. This yields three key findings. \textbf{First}, DMs struggle to correctly generate elements that are not explicitly named in the prompt. \textbf{Second}, we find the prompt embedding contains disentangled representations for texture and physics, which enables targeted causal interventions. \textbf{Third}, we localize the emergence of visual causal structure to the initial, computationally limited denoising steps.

Based on these findings, we introduce \textbf{\LINA (Learning INterventions Adaptively)}. Leveraging the pre-trained DM's intrinsic world knowledge without fine-tuning, \LINA applies adaptive guidance to enforce causal consistency, satisfying both explicit instructions (OOD scenarios) and implicit physical laws. \LINA performs (1) targeted guidance in the denoising process, governed by a lightweight \textbf{Adaptive Intervention Module (AIM)}, which is trained to predict the intervention strength, and (2) a reallocated, causality-aware denoising schedule. Our method outperforms SOTA on challenging causal generation tasks in our PAP dataset and Winoground dataset, without relying on external Multimodal Large Language Models (MLLMs) \citep{wan2025maestro,lv2025multimodal,huang2024smartedit} during inference.

Overall, our contributions are as follows:
\begin{enumerate}[itemsep=0em,label={(\arabic*)},leftmargin=1cm]
\item We introduce the \textbf{Causal Scene Graph (CSG)}, a representation that unifies causal dependencies and spatial layouts, providing a basis for diagnostic interventions.
\item We construct the \textbf{Physical Alignment Probe (PAP)}, a dataset consisting of structured prompts, SOTA-generated images, and fine-grained masks to quantify DMs' physical and OOD failures.
\item We perform a \textbf{diagnostic analysis} via CSG-guided masked inpainting, providing the first quantitative evaluation of DMs' multi-hop reasoning failures through bidirectional probing of edges in the CSG.
\item We propose \textbf{\LINA}, a framework that learns to predict and apply prompt-specific guidance, achieving SOTA alignment on image and video DMs without MLLM inference or retraining.
\end{enumerate}
\section{Preliminaries}
\label{sec:Preliminaries}

\begin{figure}[t]
    \centering
    \includegraphics[width=0.9\columnwidth]{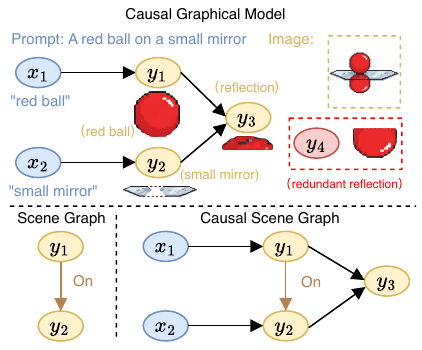}
    \caption{\textbf{Examples of CGM, SG, and CSG.} (1) CGM depicts the causal structure of the prompt and its corresponding direct or indirect elements in the generated image. We segment the prompt $\textcolor{myblue}{X}$ and the generated image $\textcolor{myyellow}{Y}$ into the semantic units $\{
\textcolor{myblue}{x_1}, \textcolor{myblue}{x_2}\}
$ and $
\{
\textcolor{myyellow}{y_1}, 
\textcolor{myyellow}{y_2}, 
\textcolor{myyellow}{y_3}, 
\textcolor{myred}{y_4}
\}$.
$\textcolor{myblue}{x_1}$ and $\textcolor{myblue}{x_2}$ in the prompt directly cause $\textcolor{myyellow}{y_1}$ and $\textcolor{myyellow}{y_2}$. $\textcolor{myyellow}{y_3}$ arises from the physical interaction between $\textcolor{myyellow}{y_1}$ and $\textcolor{myyellow}{y_2}$. $\textcolor{myred}{y_4}$ is recognized as redundant since it does not conform to causal relationship.
(2) SG depicts the spatial relationship specified by the prompt. \eg, $\textcolor{myyellow}{y_1}$ ``on'' $\textcolor{myyellow}{y_2}$. (3) CSG reuses nodes from both CGM and SG, allowing for a complete representation of dependencies in a single graph and enabling interventions.}
    \label{fig:CSG}
\end{figure}

\subsection{Causal Graphical Model}
A Causal Graphical Model (CGM) captures the causal structure among variables in a system \citep{peters2017elements, DBLP:conf/iclr/YuL25}. It comprises a set of variables $X_1, \ldots, X_n$ along with a joint distribution, represented as a directed acyclic graph (DAG). In this graph, each node corresponds to a variable, and each directed edge $X_i \rightarrow X_j$ denotes a direct causal influence from $X_i$ to $X_j$.
CGMs can be used to describe causal relationships in image generation as shown in Fig. \ref{fig:CSG} (a). We denote the prompt as $X$ and the generated image as $Y$. Both $X$ and $Y$ are partitioned into semantic units, represented as $X=\{x_1, x_2, \ldots\}$ and $Y=\{y_1, y_2, \ldots\}$.
We define: 

\textit{\textbf{Direct elements $\boldsymbol{Y_D}$}}: entities $y_i \in Y$ that are exactly one directed edge away from some $x_j \in X$ in the CGM, \ie, elements explicitly determined in the prompt. 

\textit{\textbf{Indirect elements $\boldsymbol{Y_I}$}}: entities $y_k \in Y$ that are causally downstream of some $x_j$ but separated by one or more intermediate nodes, often arising from implicit physical laws (\eg, shadows, reflections). 
Causality in DMs has a clear hierarchy: $\boldsymbol{X \rightarrow Y_D \rightarrow Y_I}$. We will adhere to these definitions throughout the paper.

\subsection{Scene Graph}

A Scene Graph (SG) is another form of structured image representation, organizing an image into elements (nodes) and their pairwise spatial relationships (edges) \citep{chang2021comprehensive}. Unlike CGMs, SG may contain cycles (\eg, three people each ``near'' one another).
We define:

\textit{\textbf{Direct layout}}: spatial relations explicitly stated in the prompt (\eg, ``a lightbulb surrounding some plants''). Such relations can appear in OOD instructions.

\textit{\textbf{Indirect layout}}: spatial relations not stated in the prompt but implied by physical laws (\eg, density implies ``iron under water surface''). We will adhere to these definitions throughout the paper.
\section{Causal Diagnostics of Diffusion Models}
\label{sec:diagnostics}

In this section, we diagnose the root of DMs' causal failures. We first introduce the Causal Scene Graph (CSG) to formally define the problem. Then we introduce the Physical Alignment Probe (PAP) dataset to quantify these failures. Finally, we use PAP to conduct two sets of probe experiments to localize the failure point (indirect elements in multi-hop reasoning) and the failure source (guidance miscalibration in the generation process).

\begin{figure}[t]
    \centering
    \includegraphics[width=0.7\columnwidth]{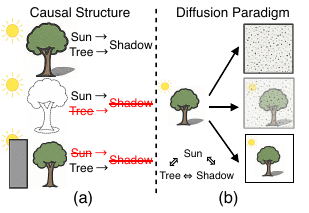}
    \caption{\textbf{Limitation of DMs' learning paradigm.}
    (a) In reality, removing a tree causes its shadow to disappear, whereas obscuring the shadow does not remove the tree. The causal relationship has a \textbf{clear direction}. (b) In DMs, both the tree and its shadow are learned to be denoised \textbf{simultaneously} without causal direction.
    }
    \label{fig:DM_limitation}
\end{figure}

\subsection{Causal Scene Graph and Problem Formulation}
\label{sec:csg_and_formulation}
We introduce the Causal Scene Graph (CSG). Formally, we define a CSG as a graph $G_{\text{CSG}} = (V, E)$, where the node set $V = X \cup Y$ comprises all semantic units from the prompt $X=\{x_i\}$ and the image $Y=\{y_j\}$. The edge set $E = E_C \cup E_S$ is the union of two sets. $E_C$ contains \textbf{Causal Edges} ($v_i \rightarrow v_j$), which form a DAG to model the directional flow of physical influence. $E_S$ contains \textbf{Spatial Edges}, which model the spatial layout of elements.

\begin{figure}[t]
\centering
\includegraphics[width=0.95\columnwidth]{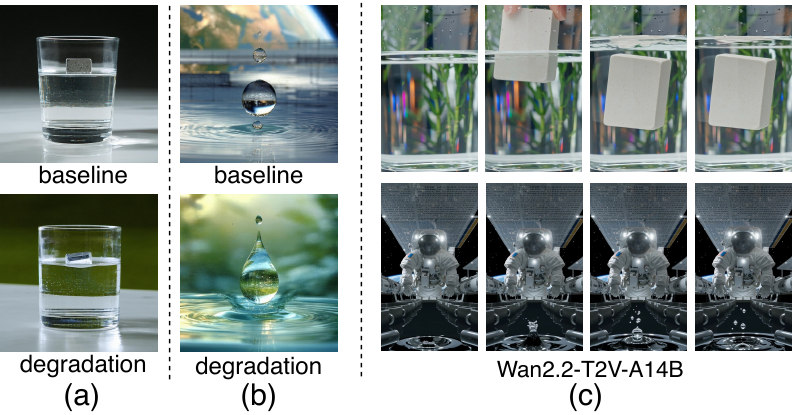}
\caption{\textbf{Multi-hop reasoning failures in DMs.}
(a) Generated with ``a small iron block in a glass full of water,'' the iron is floating. After intervention of the ``iron'' token, the iron degrades into an ice cube. The DM is merely performing texture replacement on common layouts.
(b) Generated with ``a water drop in the space station,'' the water surface is inconsistent with microgravity. After intervention, it degrades into a generic water surface.
(c) Results from the video DM Wan-2.2 \citep{wan2025wan} using the above prompts. The texture misalignment occurs even without intervention.
}
\label{fig:reasoning_failures}
\end{figure}

This directional representation highlights an inherent limitation in DM's paradigm. In conventional DM training, all elements within an image are denoised simultaneously. Consequently, causally hierarchical elements (\eg, a tree as the cause $Y_D$ and its shadow as the effect $Y_I$) are learned concurrently and symmetrically in a \textbf{flattened, non-hierarchical} paradigm, as illustrated in Fig.~\ref{fig:DM_limitation}. While data scaling has mitigated this \citep{esser2024scaling}, SOTA DMs still primarily capture texture correlations, as shown in Fig.~\ref{fig:reasoning_failures}.

We can formalize this problem using CSG. Let $X$ be the user's text prompt, which implies a ground-truth \textbf{Target CSG}, $G_X^*$. Specifically, a DM employs a denoising network $\epsilon_\theta$ in the iterative sampling dynamics, defining the probabilistic mapping $\mathcal{F}$ that generates an image $Y$. From this image, we can extract the \textbf{Generated CSG}, $G_X^{\text{gen}} = \mathcal{G}(Y)$, by parsing its underlying semantic elements and structural relationships.

Ideally, the mapping $\mathcal{F}$ should perfectly reproduce $G_X^*$. However, our diagnostics will show that this mapping $\mathcal{F}$ is \textbf{distorted}. The DM's internal knowledge is applied inconsistently, leading to physical misalignments and OOD failures, such that the sampling probability $P(G_X^{\text{gen}} = G_X^*) \ll 1$ for hard cases. CSG provides the theoretical basis and interventional tools to diagnose this distortion.

\subsection{The Physical Alignment Probe (PAP) Dataset}
\label{sec:pap_dataset}
To quantitatively measure the distortion of the mapping $\mathcal{F}$, we construct the \textbf{Physical Alignment Probe (PAP) dataset}. PAP is a multi-modal corpus based on CSG for diagnosing physical reasoning and OOD generation. It comprises three core components: (1) a structured library of prompts, (2) a large-scale image corpus generated from these prompts using SOTA DMs (SD-3.5-large \citep{esser2024scaling} and FLUX.1-Krea-dev \citep{FLUX.1-Krea-dev}), and (3) fine-grained segmentation masks to facilitate diagnostic interventions.

Specifically, the prompt library consists of 287 prompts, which are divided into three diagnostic subsets: \textbf{Optics} and \textbf{Density} \citep{han2025can,meng2024phybench}, which probe adherence to implicit physical laws, and a dedicated \textbf{OOD} subset focusing on counterfactual attributes and spatial relations.
We employ an MLLM-based evaluator (Qwen2.5-VL-72B \citep{bai2025qwen2}) with rule-based checks to parse the generated CSG ($G_X^{\text{gen}}$) and compare it against the prompt's ground-truth ($G_X^*$).
To quantify failures, we define two metrics based on our CGM definitions (Sec.~\ref{sec:Preliminaries}):
(1) \textbf{Texture Alignment}: The success rate for generating \textit{direct elements} ($\boldsymbol{Y_D}$).
(2) \textbf{Physical Alignment}: The success rate for generating \textit{indirect elements} ($\boldsymbol{Y_I}$).

We generate 50 images per prompt for each DM, forming the 28,700-image corpus that serves as the basis for our baseline evaluation. The MLLM evaluator provides bounding boxes and point prompts for elements within these images, which we utilize to generate precise segmentation masks via SAM2 \citep{ravi2024sam}.

\subsection{Probe DM's Multi-Hop Reasoning}
\label{sec:probe_inpainting}
To systematically diagnose the failure in the causal hierarchy $X \rightarrow Y_D \rightarrow Y_I$, we use the masks from our PAP dataset to conduct a series of masked inpainting probes. Adopting the notations from Sec.~\ref{sec:Preliminaries}, we design five diagnostic experiments to inspect the causal edges bidirectionally, aiming to distinguish between robust directional reasoning and superficial symmetric correlations:

($\mathrm{I}$) $P(Y_I|X, Y_D)$, \textbf{forward inference}. The success rate (81.1\%) shows no significant improvement over the baseline (80.4\%).
\textbf{Finding:} DMs fail to leverage visual context ($Y_D$) to improve physical reasoning for indirect elements.

($\mathrm{II}$) $P(Y_D|X, Y_I)$, \textbf{mediator inference}. The success rate (94.9\%) is similar to the baseline (94.3\%).

($\mathrm{III}$) $P(Y_D|X, Y_I')$, \textbf{mediator inference with conflict}, where $Y_I'$ conflicts with $X$ (\eg, $X$ specifies a ``\textcolor{myred}{red} ball'' but $Y_I'$ is a ``\textcolor{mygreen}{green} reflection''). In 94.0\% of cases, the generated $Y_D$ aligns with $X$, while 0.4\% aligns with the conflicting $Y_I'$.
\textbf{Finding:} Experiments $\mathrm{II}$ and $\mathrm{III}$ show that DMs prioritize direct supervision from $X$ when generating $Y_D$, largely ignoring downstream causal consistency with $Y_I$.

($\mathrm{IV}$) $P(Y_I|Y_D)$, \textbf{forward inference without prompt}. The success rate plummets to 2.2\%.

($\mathrm{V}$) $P(Y_D|Y_I)$, \textbf{backward inference without prompt}. The success rate is 2.4\%.
\textbf{Finding:} Experiments $\mathrm{IV}$ and $\mathrm{V}$ demonstrate that the prompt $X$ is essential for establishing the causal relationship; visual context alone is insufficient.

Collectively, these probe results uncover a structural deficiency in DMs' reasoning: the generation of the effect $Y_I$ is not conditionally dependent on the visual state of the cause $Y_D$. Consequently, the causal hierarchy $X \rightarrow Y_D \rightarrow Y_I$ collapses into a parallelized, correlation-based mapping $(X \rightarrow Y_D, X \rightarrow Y_I)$, where the requisite physical consistency between visual elements is neglected.

\begin{figure}[t]
 \centering
 \includegraphics[width=\columnwidth]{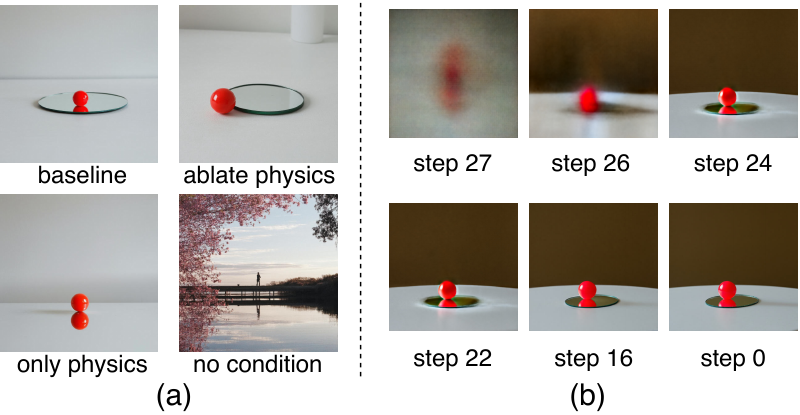}
 \caption{
 \textbf{(a)} The prompt embedding contains decomposable representations for textures and physical relations. Ablating relation tokens (\eg, ``on'') removes spatial layout and physical interactions, while retaining only these tokens produces the core physical texture but loses object semantics.
\textbf{(b)} The denoising process establishes the causal structure in the initial, computationally limited steps (step 26 to 24).
}
\label{fig:probing_causality}
\end{figure}

\subsection{Causal Representation in Embedding Space}
\label{sec:probe_embedding}
Having localized the causal failure in multi-hop reasoning, we investigate its representational basis within the prompt embedding space. Specifically, we examine the dependence of Physical and Texture Alignment on different tokens.

First, we perform intervention on \textbf{relation tokens}. We define them as the linguistic units governing the topological structure of the CSG. Specifically, they encompass (1) \textbf{interaction verbs} (\eg, ``hits'', ``eats'') that explicitly denote causal actions ($E_C$), and (2) \textbf{spatial prepositions} (\eg, ``on'', ``in'') that define the spatial configuration ($E_S$), establishing the requisite conditions for implicit physical interactions ($E_C$, \eg, reflections). We replace these tokens with the ones at the same indices in the unconditional embedding tensor. The Physical Alignment success rate collapses from its \textbf{80.4\%} baseline to just \textbf{5.2\%}, while Texture Alignment remains high (\textbf{91.8\%} vs. its \textbf{94.3\%} baseline).

Second, we perform intervention on \textbf{object and attribute tokens} (\eg, ``red'', ``ball''). As a result, the Texture Alignment collapses to \textbf{3.1\%}. Yet, Physical Alignment is remarkably preserved (\textbf{92.5\%}), as the model still generates the \textbf{causal phenomenon} (\eg, a reflection texture) even without the specific object. An illustrative example of this disentanglement is shown in Fig.~\ref{fig:probing_causality} (a).

These findings fundamentally reframe the physical misalignment: it is not a knowledge deficit within the denoising network $\epsilon_\theta$, but a \textbf{miscalibrated guidance signal} driving the mapping $\mathcal{F}$.
Given that the extent of miscalibration varies across different targets $G_X^*$, we can effectively steer $\mathcal{F}$ toward $G_X^*$ by \textbf{adaptively} modulating the influence of relation tokens, without altering the underlying model $\epsilon_\theta$.

\begin{figure*}[t]
\centering
\includegraphics[width=1\textwidth]{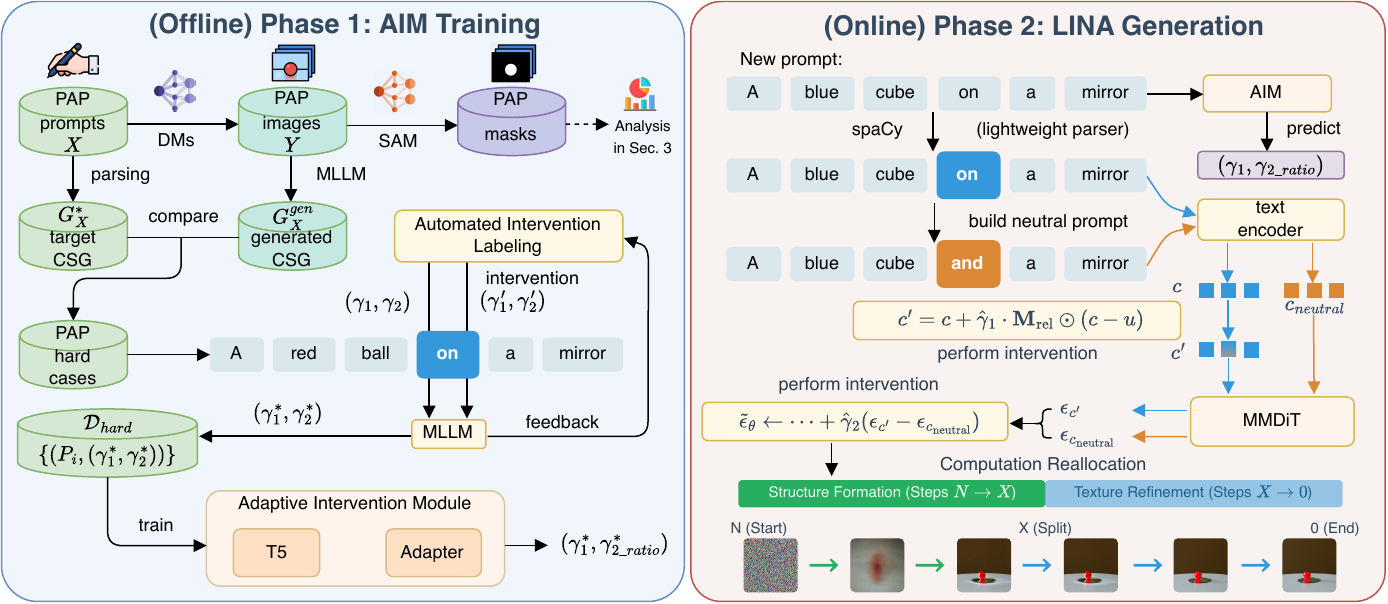}
\caption{
\textbf{Overview of the \LINA framework, which operates in two phases.}
\textbf{(Left) Phase 1: (Offline) AIM Training.} We identify baseline failures (``hard cases'') from our PAP dataset. An MLLM evaluator performs an automated comparative search to find optimal intervention strengths $(\gamma_1^*, \gamma_2^*)$ for these prompts. This creates a dataset $\mathcal{D}_{hard}$, which is used to train the \textbf{Adaptive Intervention Module (AIM)} to predict these strengths directly from a prompt.
\textbf{(Right) Phase 2: (Online) \LINA-Guided Generation.} For a new prompt $X_{new}$, the pre-trained AIM predicts the $(\hat{\gamma}_1, \hat{\gamma}_{2\_ratio})$. \LINA consists of three components: (1) A \textbf{Token-Embedding-level Intervention ($\gamma_1$)} enhances relation tokens (Eq.~\ref{eq:c_prime}), (2) a \textbf{Visual-Latent-level Intervention ($\gamma_2$)} introduces a contrastive guidance term (Eq.~\ref{eq:full_guidance}), and (3) \textbf{Computation Reallocation} concentrates the inference budget on the initial \textbf{Structure Formation} phase to prioritize the establishment of causal structure.
}
\label{fig:method_overview}
\end{figure*}

Furthermore, we analyze the denoising process to pinpoint \textit{when} this causal structure emerges. We inspect the intermediate latent states from SD-3.5-large's official 28-step denoising schedule. As shown in Fig.~\ref{fig:probing_causality} (b), the causal structure is disproportionately established at the very beginning of the process. In \textbf{97.8\%} of successful generations on the PAP-Optics subset, the correct structure is identifiable within the initial 2-4 iterations (\ie, at steps 26-24 of the 28-step reverse denoising schedule, corresponding to the highest noise levels).
This finding shows that the causal structure is set in these early, computationally limited steps, while subsequent steps primarily refine texture. This strongly motivates a computational reallocation on this critical \textbf{causal structure formation} period.
\section{Method}
\label{sec:Method}

In this section, we introduce \textbf{\LINA (Learning INterventions Adaptively)}.
Based on our diagnostics in Sec.~\ref{sec:probe_embedding}, we formalize two targeted intervention mechanisms (Sec.~\ref{sec:method_intervention}): \textbf{Token-Embedding-level} Causal Edge Calibration and \textbf{Visual-Latent-level} Contrastive Causal Guidance.
Leveraging these mechanisms, \LINA adaptively calibrates the sampling dynamics of $\mathcal{F}$ according to different input prompts (target $G_X^*$), as shown in Fig.~\ref{fig:method_overview}.
First, in the offline phase (Sec.~\ref{sec:method_aim_training}), we utilize an MLLM as a robust evaluator to guide an automated search for optimal intervention strengths. These discovered optima serve as supervision to train a lightweight \textbf{Adaptive Intervention Module (AIM)}.
Second, in the online phase (Sec.~\ref{sec:method_online_generation}), the intervention strengths predicted by the AIM are applied at inference time in conjunction with a \textbf{Causality-Aware Denoising Schedule}, which reallocates computation to the critical causal structure formative period. This process operates without MLLM overhead or DM retraining.

\subsection{Causal Intervention Mechanisms}
\label{sec:method_intervention}

\paragraph{Token-Embedding-level: Causal Edge Calibration ($\gamma_1$)}
We formulate $\gamma_1$ to calibrate the causal edge $Y_D \rightarrow Y_I$. This mechanism generalizes the token replacement intervention used in our diagnostics (Sec.~\ref{sec:probe_embedding}) to a continuous, \textit{soft} guidance strength. To implement this, we identify relation tokens via a lightweight language parser (spaCy \citep{Honnibal_spaCy_Industrial-strength_Natural_2020}) and construct a binary mask $\mathbf{M}_{\text{rel}}$ targeting their embeddings.
Given a conditional embedding $c = Emb(X)$ and an unconditional embedding $u = Emb(\varnothing)$, we compute the calibrated embedding $c'$:
\begin{equation}
\label{eq:c_prime}
c' = c + \gamma_1 \cdot \mathbf{M}_{\text{rel}} \odot (c - u)
\end{equation}
where $\odot$ denotes element-wise multiplication. By scaling the semantic direction vector $(c - u)$, this intervention selectively amplifies or attenuates the causal signal strength within the prompt embedding prior to the denoising process.

\begin{figure*}[t]
    \centering
    \includegraphics[width=\linewidth]{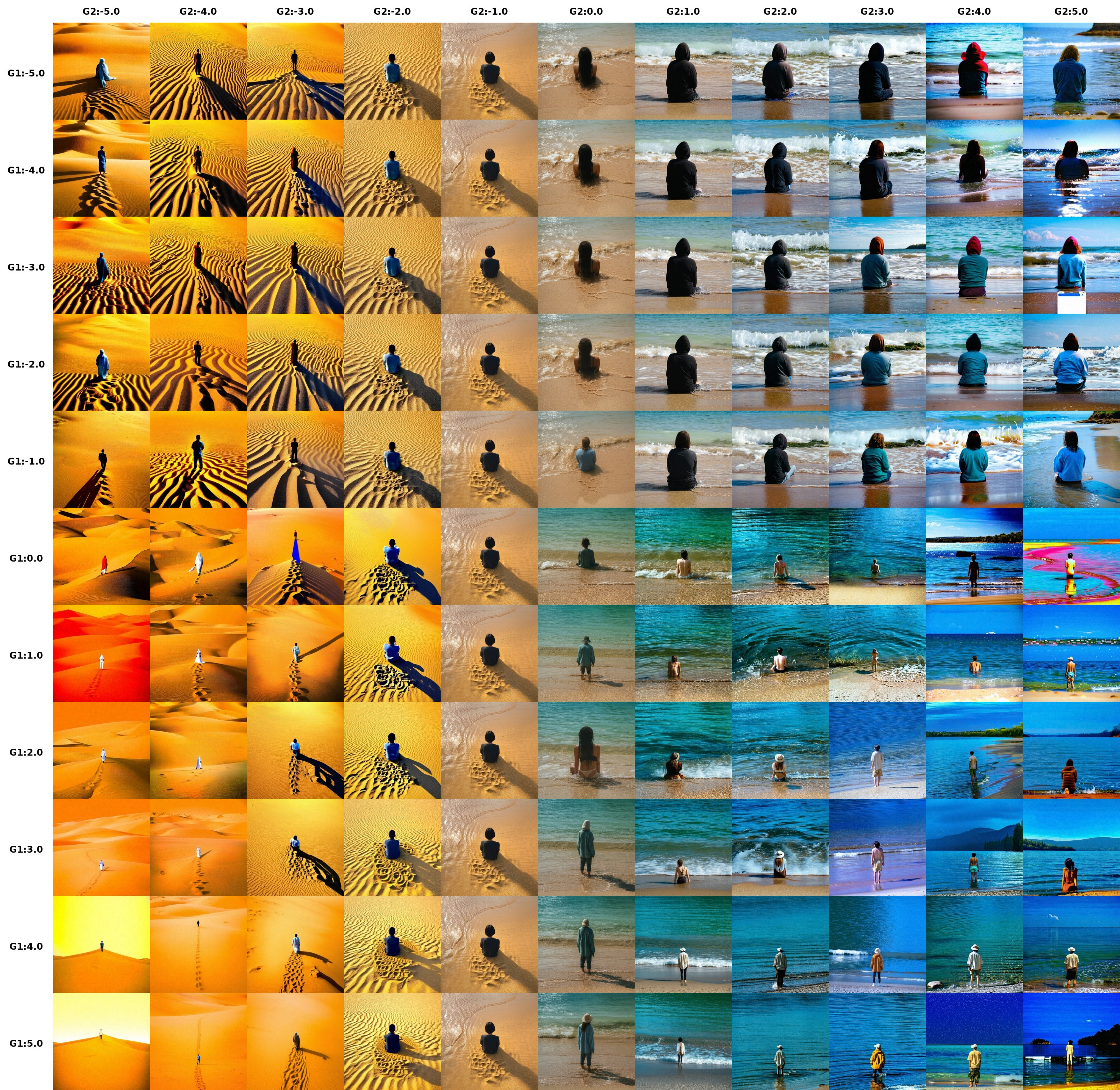}
    \caption{\textbf{Optimization Landscape Analysis.} We perform a grid search for the prompt ``a person is \textit{close to} the water and \textit{in} the sand''. The effects are largely \textbf{disentangled and monotonic}: $\gamma_2$ (horizontal) controls texture (water vs. sand), while $\gamma_1$ (vertical) calibrates spatial layout and interactions. These findings validate the efficacy of our coordinate descent strategy.}
    \label{fig:grid_search_landscape}
\end{figure*}

\paragraph{Visual-Latent-level: Contrastive Causal Guidance ($\gamma_2$)}
We introduce $\gamma_2$ to enforce structural constraints by employing a targeted contrastive guidance mechanism. This formulation aims to isolate the causal signal by contrasting the target prompt against a neutral baseline, effectively filtering out the generic semantic information to accentuate the specific causal dependencies.
To achieve this, we automatically construct a ``neutral'' reference prompt $X_{\text{neutral}}$ by relaxing the specific relation tokens in $X$ to generic conjunctions (\eg, substituting ``on'' with ``and''). The guidance term derived from $X_{\text{neutral}}$ serves as a negative baseline, representing the generation of entities without their requisite causal dependencies.

The final noise prediction $\tilde{\epsilon}_{\theta}$ at timestep $t$ is computed by combining the standard Classifier-Free Guidance (CFG) \citep{ho2022classifier} with our contrastive term:
\begin{equation}
\label{eq:full_guidance}
\begin{split}
\tilde{\epsilon}_{\theta}(x_t, t) = & \ \epsilon_{\theta}(x_t, u, t) \\
& + \gamma_0 (\epsilon_{\theta}(x_t, c', t) - \epsilon_{\theta}(x_t, u, t)) \\
& + \gamma_2 (\epsilon_{\theta}(x_t, c', t) - \epsilon_{\theta}(x_t, c_{\text{neutral}}, t))
\end{split}
\end{equation}
where $c'$ is the calibrated embedding from Eq.~\ref{eq:c_prime}, and $c_{\text{neutral}} = Emb(X_{\text{neutral}})$. By subtracting the neutral component, the $\gamma_2$ term explicitly calibrates the guidance signal responsible for the specific spatial and physical interactions defined in $X$.

\subsection{Phase 1: AIM Training}
\label{sec:method_aim_training}

The offline phase aims to train an adaptive mapping $\Phi$, which predicts intervention strengths $\Phi(X)$ conditioned on the input prompt $X$ to ensure causal alignment. To construct the training dataset $\mathcal{D}_{hard}$, we identify baseline failures (hard cases) from the PAP image corpus (Sec.~\ref{sec:pap_dataset}) and determine their optimal intervention strengths $(\gamma_1^*, \gamma_2^*)$ via an MLLM-based automated search.

\paragraph{Automated Intervention Labeling.}
Manually calibrating the target intervention strengths $(\gamma_1, \gamma_2)$ for every hard case input is intractable. To address this, we introduce an automated labeling pipeline driven by our MLLM evaluator. This pipeline identifies the optimal intervention strengths $(\gamma_1^*, \gamma_2^*)$ for each hard case $X_i$, supported by an analysis of the intervention landscape.

\begin{figure}[t]
    \centering
    \includegraphics[width=0.8\linewidth]{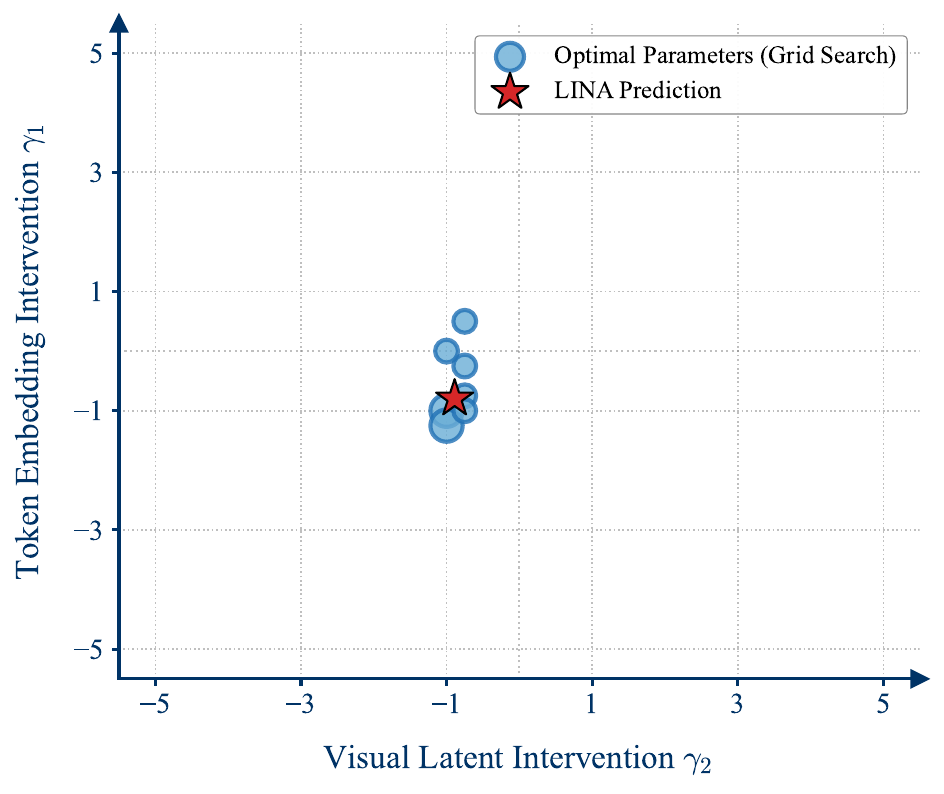}
    \caption{\textbf{Stability of Optimal Interventions.} For a same prompt, we plot the optimal $(\gamma_1, \gamma_2)$ strengths for failure cases across 100 random seeds (blue dots). The high density of the cluster indicates that the requisite correction is a stable, structural property of the prompt. The \LINA prediction (red star) aligns with the cluster center.}
    \label{fig:scatter_stability}
\end{figure}

\begin{enumerate}[itemsep=0em,label={(\arabic*)}]
\item \textbf{Identify Baseline Failures:} We identify the prompts and seeds from the PAP dataset where baseline DMs fail to satisfy the causal requirements.

\item \textbf{Coordinate Descent Search:} Given the high computational cost of the generation-evaluation loop, we employ an efficient Coordinate Descent strategy. As visualized in Fig.~\ref{fig:grid_search_landscape}, we observe that the generative responses to Visual-Latent-level ($\gamma_2$) and Token-Embedding-level ($\gamma_1$) interventions are largely \textbf{disentangled and monotonic}. These findings validate the efficacy of our coordinate descent strategy.

\item \textbf{Construct Target Dataset:} This process yields our structured training dataset, $\mathcal{D}_{hard} = \{ (X_i, (\gamma_1^*, \gamma_2^*)) \}_{i=1}^N$. Crucially, as shown in Fig.~\ref{fig:scatter_stability}, we observe that the optimal intervention strengths are \textbf{robust to initialization noise}, clustering tightly for a given prompt across random seeds. This stability confirms the feasibility of training a deterministic mapping $\Phi(X)$ to predict these values.
\end{enumerate}

\paragraph{AIM Training.}
We design the \textbf{Adaptive Intervention Module (AIM)}, denoted as $\Phi$, to generalize this knowledge. $\Phi$ consists of a pre-trained T5 text encoder \citep{raffel2020exploring} followed by a lightweight MLP regression head. It is trained on $\mathcal{D}_{hard}$ using a standard L2 regression loss to predict the intervention strengths:
\begin{equation}
\mathcal{L} = \sum_{i=1}^N \| \Phi(X_i) - (\gamma_1^*, \gamma_2^*/\gamma_0) \|_2^2
\end{equation}
We predict the ratio $\gamma_2^*/\gamma_0$ (as $\hat{\gamma}_{2\_ratio}$) for a normalized and more stable regression.

\subsection{Phase 2: Causality-Aware Denoising}
\label{sec:method_online_generation}

In the online generation phase, the pre-trained AIM is deployed to guide the generative process for any new user prompt $X_{new}$.

\paragraph{Pre-processing}
For a new prompt $X_{new}$, we first compute the embeddings $u = Emb(\varnothing)$, $c = Emb(X_{new})$, and $c_{\text{neutral}} = Emb(X_{\text{neutral}})$. Simultaneously, the AIM predicts the intervention strengths $(\hat{\gamma}_1, \hat{\gamma}_{2\_ratio}) = \Phi(X_{new})$, and the parser identifies the relation token mask $\mathbf{M}_{\text{rel}}$. The final latent-level strength is set as $\hat{\gamma}_2 = \hat{\gamma}_{2\_ratio} \cdot \gamma_0$.

\paragraph{Causality-Aware Denoising Schedule}
Motivated by our finding (Sec.~\ref{sec:probe_embedding}) that the visual causal structure \textit{emerges} predominantly during the initial high-noise phase, we implement a Computation Reallocation strategy to prioritize this critical period. 
We build upon the time-shifting parameterization employed in modern DMs (\eg, SD-3.5-large \citep{esser2024scaling}), which maps the standard uniform time steps $\tau \in [0, 1]$ to a shifted schedule $\tau_s$:
\begin{equation}
    \tau_s = \frac{s \cdot \tau}{1 + (s - 1) \cdot \tau}
\end{equation}
where $s$ is the shift parameter. 
While SOTA models typically set $s > 1$ to accommodate high resolutions, we identify that this default shift is often insufficient for complex causal reasoning. 
Therefore, we explicitly \textbf{recalibrate} $s$ by amplifying the shift, thereby further concentrating the sampling density in the early structure-formation phase. This adjustment ensures that a larger proportion of the inference budget is dedicated to establishing a robust causal layout before texture refinement begins. The full \LINA-guided noise prediction $\tilde{\epsilon}_{\theta}$ (Eq.~\ref{eq:full_guidance}) is applied throughout this causality-aware schedule.
\section{Experiments}
\label{sec:experiments}

\subsection{Experimental Setup}
\label{sec:exp_setup}
We conduct experiments using two SOTA DMs: SD-3.5-large \citep{esser2024scaling} and FLUX.1-Krea-dev \citep{FLUX.1-Krea-dev}. We employ Qwen2.5-VL-72B \citep{bai2025qwen2} as our MLLM evaluator for automated evaluation, following the methodology described in Sec.~\ref{sec:pap_dataset}.

\subsection{Baselines}
\label{sec:baselines}
Our baselines include
(1) \textbf{SOTA Closed-Source Models}: Nano Banana \citep{Gemini-image} and GPT-Image \citep{GPT-Image}. These serve as black-box baselines and editing systems.
(2) \textbf{Top-tier Open-Source DMs}: SD-3.5-large (SD-3.5) \citep{esser2024scaling}, FLUX.1-Krea-dev (FLUX.1) \citep{FLUX.1-Krea-dev}, and Wan2.2-T2V-A14B (Wan-2.2) \citep{wan2025wan}. These models represent the baseline mapping $\mathcal{F}$ that we apply corrections to.

We also select three open-source methods representing the three primary correction strategies.
(3) \textbf{Prompt Engineering Methods}: LLM-grounded Diffusion (LMD) \citep{lian2023llm}. This strategy corrects the \textit{input} to $\mathcal{F}$, seeking $X'$ such that $\mathcal{F}(G_{X'}) \approx G_X^*$.
(4) \textbf{MLLM-in-the-loop Methods}: Ping-Pong-Ahead Diffusion (PPAD) \citep{lv2025multimodal}. This strategy uses an external corrector $M$ to guide the \textit{output} of $\mathcal{F}$, achieving $M(\mathcal{F}(G_X^*)) \approx G_X^*$.
(5) \textbf{Finetuning Methods}: LoRA finetuning implemented in Diffusers \citep{von_Platen_Diffusers_State-of-the-art_diffusion}. This strategy modifies the mapping $\mathcal{F}$ by retraining the DM's weights $\theta$.

\subsection{Main Results}

\begin{figure}[t]
\centering
\includegraphics[width=0.87\columnwidth]{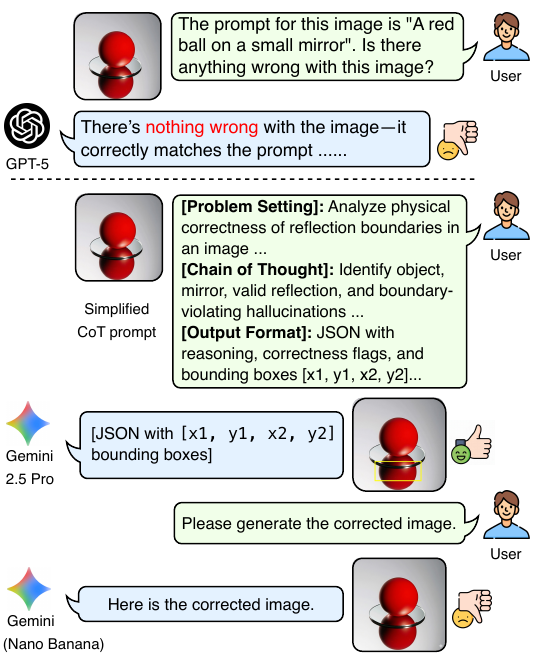}
\caption{\textbf{Gap between Causal Identification and Generative Correction in SOTA systems.} (Top) GPT-5 \citep{GPT-5} fails to recognize a physical violation (redundant reflection). (Bottom) With a detailed, chain-of-thought \citep{wei2022chain} style prompt, Gemini 2.5 Pro \citep{comanici2025gemini} successfully performs discrimination by identifying and localizing the error. However, the subsequent correction by Nano Banana \citep{Gemini-image} still produces a physically flawed image.}
\label{fig:mllm_gap}
\end{figure}

\subsubsection{Evaluation on Physical Alignment}

\paragraph{Analysis of SOTA Closed-Source Systems}
Building upon the diagnostic insights in Sec.~\ref{sec:diagnostics}, we investigate the persistent disconnect between \textit{causal identification} and \textit{generative correction}.
As exemplified in Fig.~\ref{fig:mllm_gap}, even advanced proprietary systems frequently lack the generative control to rectify the physical inconsistencies they correctly diagnose.
We quantify this ``identification-correction gap'' in Tab.~\ref{tab:editing_gap}, utilizing the $\mathcal{D}_{hard}$ test split to ensure a rigorous evaluation and API efficiency.
Our results indicate that current state-of-the-art editing models exhibit limited efficacy in resolving physical violations via standard instructions.
While integrating Chain-of-Thought (CoT) reasoning \citep{wei2022chain} enhances the intermediate diagnostic step, it fundamentally fails to bridge the translation gap to the final visual output.
In contrast, \LINA circumvents this limitation by leveraging adaptive interventions during re-generation, thereby achieving superior alignment scores (96.4\% and 86.0\%).

\begin{table}[t]
\centering
\caption{\textbf{Gap between Causal Identification and Generative Correction.} We evaluate SOTA closed-source editing models on their ability to correct physical violations from the $\mathcal{D}_{hard}$ test set. Baseline models are given the flawed image and the correction prompt. ``+ CoT'' variants perform a CoT analysis of the flaw before correction. Notably, \LINA is evaluated by re-generating from the same prompt and seed. The metric is Correction Success Rate ($\uparrow$). Best results are in \textbf{bold}.}
\label{tab:editing_gap}
\setlength{\tabcolsep}{4pt}
\begin{tabular}{lcc}
\hline
\textbf{Method} & \textbf{Optics} (\%, $\uparrow$) & \textbf{Density} (\%, $\uparrow$) \\
\hline
Nano Banana \citep{Gemini-image} & 2.5 & 25.0 \\
GPT-image \citep{GPT-Image} & 2.5 & 22.5 \\
\hline
Nano Banana + CoT & 45.0 & 70.5 \\
GPT-image + CoT & \textbf{67.5} &  \textbf{82.0} \\
\hline
\textbf{\LINA} (on SD-3.5) & \textbf{96.4} & \textbf{86.0} \\
\hline
\end{tabular}
\end{table}

\paragraph{Open-Source Methods}
All baselines are applied to SD-3.5-large for a fair comparison. As shown in Table~\ref{tab:main_physics}, \LINA significantly outperforms all baseline methods. LMD \citep{lian2023llm} enforces spatial layout but struggles to capture physical laws. PPAD \citep{lv2025multimodal} suffers from the ``identification-correction gap'' observed in the closed-source systems; its MLLM identifies flaws, but its corrections fail to override the DM's flawed internal dynamics. LoRA finetuning \citep{von_Platen_Diffusers_State-of-the-art_diffusion} overfits to correlations rather than fixing the underlying symmetric paradigm. In contrast, \LINA succeeds by diagnosing the root cause as a ``causal guidance miscalibration'' (Sec.~\ref{sec:probe_embedding}).

\begin{table}[t]
\centering
\caption{\textbf{Main Results on Physical Alignment.} We evaluate \LINA and SOTA baselines on their ability to achieve physical alignment on the \textbf{PAP-Optics} and \textbf{PAP-Density} subsets. For fair comparison, all methods (except FLUX.1) are applied on the SD-3.5-large backbone. The metric is Success Rate (\% $\uparrow$). Best results are in \textbf{bold}.}
\label{tab:main_physics}
\setlength{\tabcolsep}{4pt}
\begin{tabular}{lcc}
\hline
\textbf{Method} & \textbf{Optics} (\%, $\uparrow$) & \textbf{Density} (\%, $\uparrow$) \\
\hline
SD-3.5 (Baseline) & 80.4 & 54.2 \\
FLUX.1 (Baseline) & 86.9 & 64.3 \\
LMD (SD-3.5) \citep{lian2023llm} & 80.5 & 81.5 \\
PPAD (SD-3.5) \citep{lv2025multimodal} & 91.7 & 76.2 \\
LoRA (SD-3.5) \citep{von_Platen_Diffusers_State-of-the-art_diffusion} & 95.9 & 91.3 \\
\hline
\textbf{\LINA} (on SD-3.5) & \textbf{97.4} & \textbf{92.3} \\
\textbf{\LINA} (on FLUX.1) & \textbf{96.8} & \textbf{94.0} \\
\hline
\end{tabular}
\end{table}

\begin{figure*}[t]
    \centering
    \includegraphics[width=0.9\linewidth]{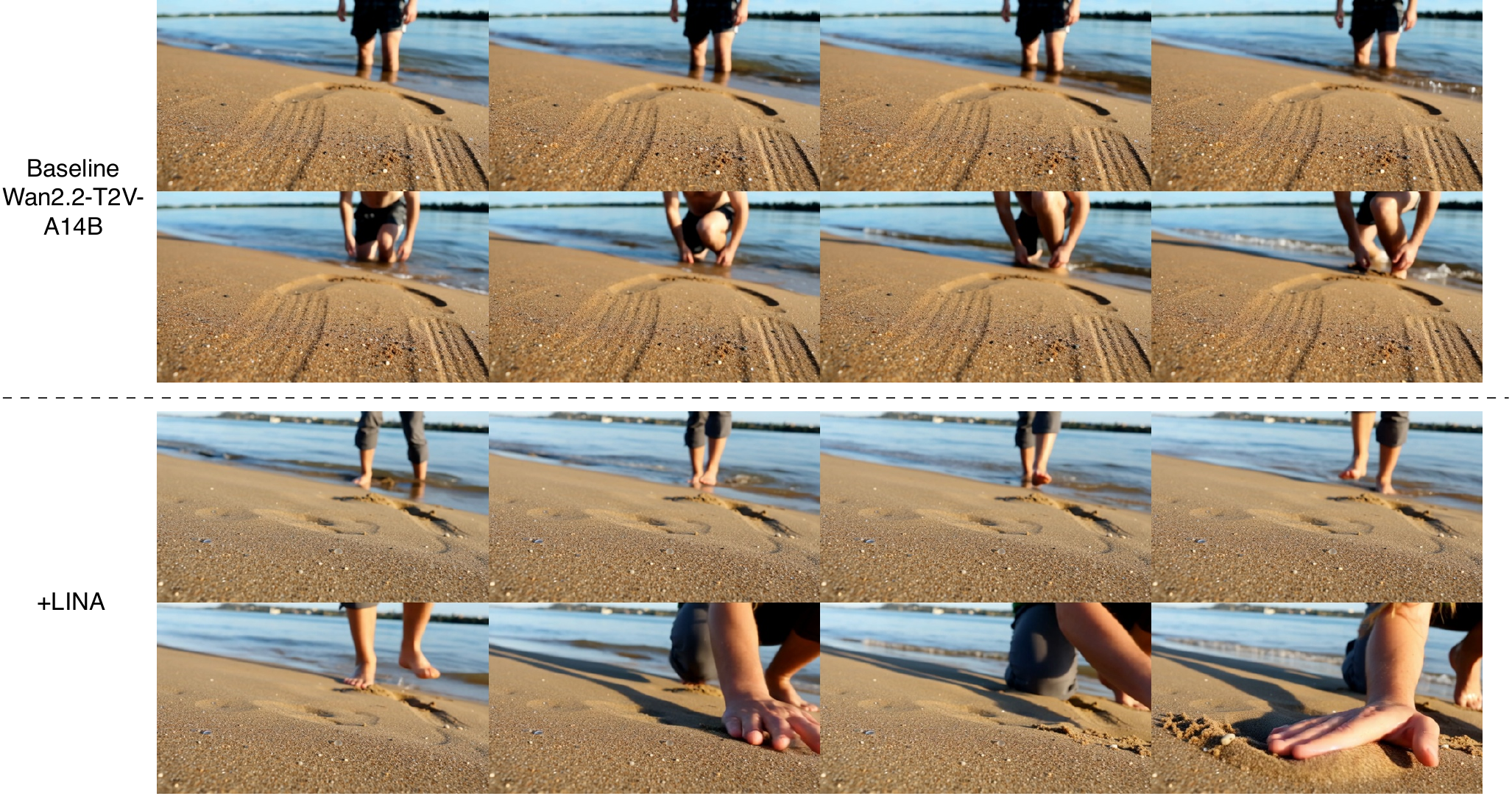}
    \caption{\textbf{Qualitative comparison on Video Generation.} We sample 8 frames uniformly from the 81-frame video. The prompt is from Winoground: ``a person is \textit{close to} the water and \textit{in} the sand''.
    \textbf{(Top) Baseline:} The model fails to capture the precise spatial preposition, incorrectly placing the person \textit{in} the water throughout the sequence.
    \textbf{(Bottom) \LINA:} Our method successfully guides the generation of a coherent temporal sequence. The person begins close to the water, moves towards the sand, and subsequently interacts with the sand (\ie, digging/entering), satisfying the complex causal and spatial requirements of the instruction.}
    \label{fig:video_example}
\end{figure*}

\subsubsection{Evaluation on OOD Instruction Following}

We evaluate OOD instruction following using two challenging benchmarks. The first is our \textbf{PAP-OOD subset}, which contains OOD spatial and causal configurations. The second is the \textbf{Winoground} \citep{thrush2022winoground} dataset. Its paired, contrastive prompts probe model biases against uncommon distributions. We adopt the 171-sample subset identified by \citet{diwan2022winoground}, which excludes ambiguous samples. As shown in Tab.~\ref{tab:main_ood}, baseline DMs fail by collapsing to prior correlations.
While LMD \citep{lian2023llm} improves performance by generating an explicit scene layout, it cannot fully override the DM's bias. Other methods, such as PPAD \citep{lv2025multimodal} and LoRA \citep{von_Platen_Diffusers_State-of-the-art_diffusion}, provide only marginal gains.
\LINA achieves SOTA performance by directly targeting this bias.

\begin{table}[t]
\centering
\caption{\textbf{Main Results on OOD Instruction Following.} We evaluate \LINA and SOTA baselines on their ability to follow OOD instructions from the Winoground (Wino.) subset and our PAP-OOD set. For fair comparison, all methods (except FLUX.1) are applied on the SD-3.5-large backbone. The metric is Success Rate (\% $\uparrow$). Best results are in \textbf{bold}.}
\label{tab:main_ood}
\setlength{\tabcolsep}{5pt}
\begin{tabular}{lcc}
\hline
\textbf{Method} & \textbf{Wino.} (\%, $\uparrow$) & \textbf{PAP-OOD} (\%, $\uparrow$) \\
\hline
SD-3.5 (Baseline) & 54.4 & 69.3 \\
FLUX.1 (Baseline) & 65.5 & 80.6 \\
+ LMD (SD-3.5) \citep{lian2023llm} & 73.1 & 75.2 \\
+ PPAD (SD-3.5) \citep{lv2025multimodal} & 62.6 & 74.1 \\
+ LoRA (SD-3.5) \citep{von_Platen_Diffusers_State-of-the-art_diffusion} & 57.3 & 72.0 \\
\hline
\textbf{\LINA} (on SD-3.5) & \textbf{79.5} & \textbf{84.3} \\
\textbf{\LINA} (on FLUX.1) & \textbf{83.0} & \textbf{86.1} \\
\hline
\end{tabular}
\end{table}

\subsubsection{Evaluation on Video Generation}
To evaluate the generalizability of \LINA, we extend our framework to the video generation domain using the SOTA model Wan-2.2 \citep{wan2025wan}.
Video generation introduces distinct challenges compared to static imaging: causality manifests as dynamic processes rather than spatial snapshots, and models exhibit heightened sensitivity to action verbs. Consequently, we utilize the \textbf{PAP-Density} subset (\eg, sinking, floating) and the \textbf{Winoground} \citep{thrush2022winoground} dataset to probe physical consistency across temporal sequences (\ie, initial state, motion dynamics, and final state), using Qwen2.5-VL-72B as the evaluator.

Quantitatively, \LINA significantly improves temporal physical alignment, boosting the success rate from a baseline of \textbf{29.5\%} to \textbf{58.0\%}.
Qualitatively, as shown in Fig.~\ref{fig:video_example}, we analyze the Winoground prompt: ``a person is \textit{close to} the water and \textit{in} the sand.'' The baseline model suffers from attribute leakage, incorrectly placing the person \textit{inside} the water. In contrast, \LINA enforces the correct causal structure, generating a temporally coherent narrative where the subject interacts with the sand while remaining adjacent to the water. This confirms that \LINA's adaptive interventions transfer effectively to the temporal domain.

\subsection{Ablation Study}
\label{sec:ablation}

We conduct an ablation study to validate \LINA's core components using SD-3.5-large. As shown in Tab.~\ref{tab:ablation}, we evaluate on PAP-Optics (Opt.), PAP-Density (Dens.), and Winoground (Wino.).

\paragraph{Effectiveness of Intervention Mechanisms}
Removing either the token-level $\gamma_1$ intervention (Eq.~\ref{eq:c_prime}) or the latent-level $\gamma_2$ intervention (Eq.~\ref{eq:full_guidance}) leads to significant performance degradation (rows 2-3). Notably, removing $\gamma_1$ (``w/o $\gamma_1$'') severely impacts OOD performance (Wino.) and complex physical reasoning (Dens.). Removing $\gamma_2$ (``w/o $\gamma_2$'') broadly undermines physical alignment across both PAP subsets. This confirms that both interventions are complementary and essential.

\paragraph{Necessity of Adaptive Intervention Module (AIM)}
We test the necessity of our adaptive AIM module by replacing it with a ``Fixed $\gamma$'' baseline, which applies a static, averaged intervention strength to all prompts. The sharp performance drop (row 4) validates our hypothesis that a prompt-specific, adaptive guidance is critical.

\paragraph{Effectiveness of Causality-Aware Schedule}
Finally, reverting to a ``Std. Schedule'' (row 5) from our reallocated, front-loaded schedule also degrades performance. This confirms our finding (Sec.~\ref{sec:probe_embedding}) that reallocating computation to the critical, early structure-formation phase is essential.

\begin{table}[t!]
\centering
\caption{\textbf{Ablation Study.} Analysis of \LINA's components on SD-3.5-large. We evaluate on PAP-Optics (Opt.), PAP-Density (Dens.), and Winoground (Wino.). Metric is Success Rate (\% $\uparrow$).}
\label{tab:ablation}
\setlength{\tabcolsep}{4pt} % Adjust column spacing for half-column
\begin{tabular}{lccc}
\hline
\textbf{Method} & \textbf{Opt. (\% $\uparrow$)} & \textbf{Dens. (\% $\uparrow$)} & \textbf{Wino. (\% $\uparrow$)} \\
\hline
\textbf{\LINA} & \textbf{97.4} & \textbf{92.3} & \textbf{79.5} \\
w/o $\gamma_1$ & 85.1 & 80.5 & 60.2 \\
w/o $\gamma_2$ & 81.3 & 78.0 & 74.9 \\
Fixed $\gamma$ & 90.5 & 85.2 & 68.4 \\
Std. Schedule & 92.3 & 88.1 & 74.3 \\
\hline
Baseline & 80.4 & 54.2 & 54.4 \\
\hline
\end{tabular}
\end{table}

\paragraph{Analysis of Evaluator Robustness}
The AIM training labels (Sec.~\ref{sec:method_aim_training}) are derived from our MLLM evaluator, Qwen2.5-VL-72B. To ensure these labels are robust and not an artifact of evaluator-specific bias, we perform an inter-evaluator agreement analysis. We compare the judgments of Qwen2.5-VL against those of GPT-5 \citep{GPT-5} and Gemini 2.5 Pro \citep{comanici2025gemini} on a 500-sample subset of the PAP dataset, measuring Krippendorff's Alpha ($\alpha$). The raw judgments from the three MLLMs achieve a high agreement ($\alpha = 0.92$). After applying our auxiliary rule-based checks to resolve simple geometric or counting errors, the agreement converges to near-perfect ($\alpha = 0.97$). This confirms that our AIM training data captures robust causal knowledge.
\section{Related Works}
\label{sec:related_works}

Our work builds on efforts to correct the causal failures of DMs. As diagnosed in Sec.~\ref{sec:csg_and_formulation}, these failures stem from a \textbf{distorted mapping} $\mathcal{F}$, wherein DM learns symmetric correlations rather than the directional causality of a target CSG ($G_X^*$). Existing approaches to remedy this distortion can be categorized by their interventional strategy.

\textbf{Prompt Engineering and Sampling Guidance.}
This path seeks a pre-distorted input $X'$ (or a modified $G_{X'}$) such that the flawed mapping $\mathcal{F}$ produces the correct output: $\mathcal{F}(G_{X'}) \approx G_X^*$. Methods include training LLMs to rewrite naive prompts \citep{wang2025promptenhancer, ji2025prompt}, or parsing prompts into structured layouts (\eg, bounding boxes) to guide attention \citep{lian2023llm, phung2024grounded, li2025ldgen}. Others use LLM reasoning to generate negative guidance \citep{hao2025enhancing}. While effective for spatial layout, these methods fundamentally fail to address implicit physical and causal laws, as they only alter the input $X$ without correcting the flawed dynamics of $\mathcal{F}$. \LINA, in contrast, intervenes directly on the dynamics by adaptively modulating guidance signals in both the prompt embedding and visual latent spaces.

\textbf{MLLM-in-the-loop Refinement.}
This path uses an external MLLM as a post-hoc corrector, $M$, which is applied iteratively to guide the generation process, achieving $M(\mathcal{F}(G_X^*)) \approx G_X^*$. This includes ``outer-loop'' systems that refine prompts between generations \citep{wan2025maestro}, and ``inner-loop'' methods that correct intermediate latents during denoising \citep{lv2025multimodal, huang2024smartedit}. These approaches suffer from prohibitive inference costs and the identification-correction gap we demonstrate in our experiments (Fig.~\ref{fig:mllm_gap}). \LINA avoids both issues by employing an MLLM evaluator \emph{only} offline to guide the search for optimal intervention strengths, which are subsequently learned by a lightweight AIM ($\Phi$), achieving adaptive calibration with zero MLLM overhead during inference.

\textbf{Finetuning and Adapter-based Methods.}
This path attempts to correct the mapping itself by modifying the DM's weights from $\theta$ to $\theta'$, thereby changing the mapping $\mathcal{F}$ to $\mathcal{F}'$. Methods include preference alignment \citep{wu2025preference, zhao2025fine} or lightweight adapters \citep{qiu2023controlling, tong2025causal}. These methods fundamentally assume a \emph{knowledge deficit} in $\theta$ that must be retrained. In contrast, our analysis (Sec.~\ref{sec:probe_embedding}) shows SOTA DMs possess physical knowledge, but it is \emph{miscalibrated}. \LINA is founded on this insight: as a training-free framework (for the DM), it does not alter the backbone weights $\theta$. Instead, our AIM $\Phi(X)$ learns to \emph{guide} the existing generative dynamics, $\mathcal{F}(\cdot ; \theta, \Phi(X))$, avoiding the high costs and catastrophic forgetting risks of finetuning.
\section{Conclusion} In this work, we introduce \LINA, a novel framework for adaptive causal intervention in diffusion models. Based on diagnostic insights from our Causal Scene Graph (CSG), \LINA learns to perform prompt-specific interventions, significantly enhancing the physical alignment and OOD instruction-following capabilities of DMs. Without relying on MLLMs during inference or requiring DM retraining, \LINA demonstrates strong generalizability across both image and video DMs. Our work sets a foundation for developing generative models that can function as robust world simulators, understanding and rendering complex causal structures.

\clearpage
\setcounter{page}{1}
\maketitlesupplementary

\appendix

\setcounter{table}{0}
\setcounter{figure}{0}

\renewcommand{\thetable}{A\arabic{table}}
\renewcommand{\thefigure}{A\arabic{figure}}

\section{The Physical Alignment Probe (PAP) Dataset}
\label{sec:appendix_dataset}

\subsection{Overview}
The Physical Alignment Probe (PAP) dataset serves as both a rigorous benchmark for evaluation and a diagnostic tool for the findings presented in Sec.~\ref{sec:diagnostics}. Grounded in the \textbf{Causal Scene Graph (CSG)} formalism, PAP is designed to quantify the distortion in the generative mapping $\mathcal{F}$ and to facilitate the training of the \textbf{Adaptive Intervention Module (AIM)}.

The dataset comprises three aligned components: (1) a systematically constructed library of text prompts $X$; (2) a large-scale corpus of corresponding images $Y$ generated by SOTA DMs; and (3) fine-grained segmentation masks that distinguish between direct elements ($Y_D$) and indirect elements ($Y_I$).
To comprehensively probe causal failures, the dataset is stratified into three distinct subsets, containing a total of 287 unique prompt templates, as detailed in Tab.~\ref{tab:pap_stats}.

\subsection{Systematic Prompt Construction}
\label{sec:appendix_prompt_construction}

The prompt library is constructed using a compositional approach (\ie, Cartesian products of attributes) to ensure comprehensive coverage. To isolate causal reasoning from environmental confounders, we append a standardized suffix (\ie, ``\texttt{, on a neutral gray background, studio lighting, high resolution}'') to the majority of prompts. The specific distribution of these prompts across different subcategories is detailed in Table~\ref{tab:pap_stats}.

\textbf{Optics Domain.}
This domain targets geometric consistency and reflection accuracy. We define components across three categories: 10 distinct colors, 5 geometric shapes (\eg, cone, tetrahedron), and 2 mirror shapes (circular, square). The Cartesian product yields 100 unique prompts following the template ``\texttt{A [color] [object] on a [mirror\_shape] mirror}''.

\textbf{Density Domain.}
This domain probes the model's adherence to implicit physical laws when the outcome state is not explicitly described. We select materials spanning a wide range of densities (\eg, cork, styrofoam, aluminum, granite, iron, platinum).
\begin{itemize}
    \item \textbf{Buoyancy:} Prompts utilize the template ``\texttt{A small [material] [shape] in a simple glass beaker of [liquid]}''. This subset includes 12 regular combinations (\eg, iron in water) and 1 special case (ice in ethanol) to test knowledge of specific chemical properties.
    \item \textbf{Balance:} Prompts utilize the template ``\texttt{A balance scale with a [mat1] cube on the left and a [mat2] cube on the right, same size, front view}''. This includes: (i) \textit{Comparison}: 10 combinations of different materials where density determines the tilt; and (ii) \textit{Control}: 5 cases of identical materials on both sides, testing the model's bias-free equilibrium generation.
\end{itemize}

\begin{table}[t]
    \centering
    \small
    \caption{\textbf{Statistics of the PAP Dataset.} The dataset consists of 287 unique prompt templates stratified into three domains. OOD scenarios account for the majority of prompts to rigorously test generalization limits against statistical priors.}
    \label{tab:pap_stats}
    \begin{tabular}{llc}
        \toprule
        \textbf{Domain} & \textbf{Subcategory} & \textbf{Count} \\
        \midrule
        \multirow{1}{*}{Optics} 
         & Reflection Consistency & 100 \\
        \midrule
        \multirow{2}{*}{Density} 
         & Buoyancy (Regular \& Special) & 13 \\
         & Balance (Comparison \& Control) & 15 \\
        \midrule
        \multirow{3}{*}{OOD} 
         & Size Reversal & 75 \\
         & Impossible Containment & 75 \\
         & Physics Violation (Buoyancy \& Balance) & 9 \\
        \midrule
        \textbf{Total} & & \textbf{287} \\
        \bottomrule
    \end{tabular}
\end{table}

\textbf{OOD Domain.}
To evaluate robustness against Out-of-Distribution scenarios, we construct prompts that violate statistical priors or physical laws.
\begin{itemize}
    \item \textbf{Size Reversal:} We invert natural size relationships between typically small (\eg, ant, coin) and large (\eg, elephant, castle) entities using the template ``\texttt{A giant [small\_real] [relation] a tiny [large\_real]}''.
    \item \textbf{Impossible Containment:} We place large objects into logically insufficient containers using the template ``\texttt{A [large\_obj] [relation] a small [container]}'' (\eg, ``\texttt{A cruise ship trapped in a glass bottle}'').
    \item \textbf{Physics Violation:} We explicitly instruct the model to generate states that contradict physical laws, creating a conflict between the text instruction and internal knowledge. This subset covers \textbf{Buoyancy Violation} (\eg, ``\texttt{An iron [shape] floating...}'') and \textbf{Balance Violation} (\eg, a scale where the lighter material is explicitly described as ``\texttt{tilted down}'').
\end{itemize}

\begin{figure}[t]
    \centering
    \includegraphics[width=1.0\linewidth]{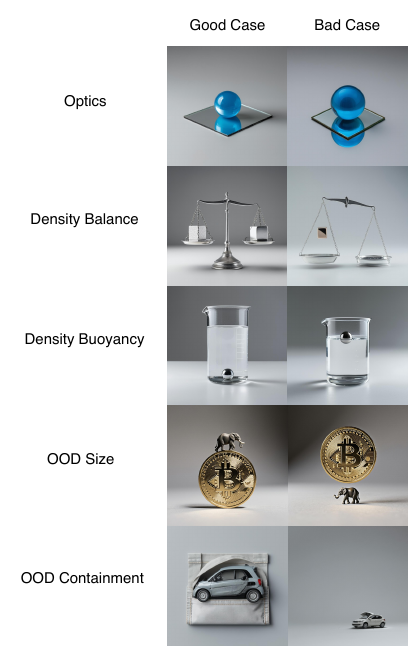}
    \caption{\textbf{Visual Examples from the PAP Dataset.} We present representative Good and Bad cases generated by DMs across the five sub-domains, highlighting the physical inconsistencies targeted by our benchmark. The specific prompts used for each row are: 
    \textbf{(1) Optics:} ``A blue ball on a square mirror, on a neutral gray background, studio lighting, high resolution''; 
    \textbf{(2) Density Balance:} ``A balance scale with an aluminum cube on the left and an aluminum cube on the right, same size, front view, on a neutral gray background, studio lighting, high resolution''; 
    \textbf{(3) Density Buoyancy:} ``A small iron ball in a simple glass beaker of water, on a neutral gray background, studio lighting, high resolution''; 
    \textbf{(4) OOD Size:} ``A giant gold coin located directly beneath a tiny elephant, on a neutral gray background, studio lighting, high resolution''; 
    \textbf{(5) OOD Containment:} ``A car inside a small pocket, on a neutral gray background, studio lighting, high resolution''.}
    \label{fig:pap_samples}
\end{figure}

\subsection{Image Corpus Generation}
\label{sec:appendix_image_corpus}

The image corpus is generated using SD-3.5-large \citep{esser2024scaling} and FLUX.1-Krea-dev \citep{FLUX.1-Krea-dev}. We adopt the official recommended settings for each model (SD-3.5: 28 steps, guidance scale 3.5; FLUX.1: 50 steps, guidance scale 4.5). To capture the variance of the generative distribution, we generate 50 unique samples for each prompt by iterating through seeds 0 to 49, resulting in a total of 14,350 images per model.

\paragraph{Annotations and Masks.}
To enable the diagnostic interventions described in Sec.~\ref{sec:diagnostics}, the PAP dataset further includes fine-grained segmentation masks and alignment labels. We derive these annotations using an automated pipeline that simultaneously performs physical verification and spatial parsing. The methodology for this evaluation and coordinate extraction is detailed in Appendix~\ref{sec:appendix_eval_pipeline}, while the specific procedure for converting these coordinates into pixel-level masks is described in Appendix~\ref{sec:appendix_mask_generation}.

\section{Automated Evaluation and Labeling Pipeline}
\label{sec:appendix_eval_pipeline}

\begin{table*}[t]
    \centering
    \small
    \renewcommand{\arraystretch}{1.2}
    \caption{\textbf{Summary of the Hybrid Evaluation Protocol.} We employ a \textit{Geometry-First} strategy. 
    \textbf{Notation:} $y^{\max}$ denotes the bottom-edge coordinate of a bounding box; $y^{\min}$ denotes the top-edge; $A$ denotes area; $\epsilon$ denotes a tolerance threshold. All geometric metrics are derived from the bounding boxes extracted by the visual parser.}
    \label{tab:eval_criteria}
    \begin{tabular}{ll p{6cm} l l}
        \toprule
        \textbf{Domain} & \textbf{Task} & \textbf{Tier 1: Geometric Constraint (Primary)} & \textbf{Threshold} & \textbf{Tier 2: Fallback} \\
        \midrule
        % --- OPTICS ---
        \multirow{2}{*}{\textbf{Optics}} & \multirow{2}{*}{Reflection} & 
        \textbf{Boundary Violation Check:} Reflection bottom ($y_{\text{refl}}^{\max}$) must not strictly exceed mirror bottom ($y_{\text{mirror}}^{\max}$). & 
        $\epsilon_{\text{spill}} = 15\text{px}$ & 
        \multirow{2}{*}{None (Strict)} \\
         & & \textit{Constraint:} $y_{\text{refl}}^{\max} \le y_{\text{mirror}}^{\max} + \epsilon_{\text{spill}}$ & & \\
        \midrule
        
        % --- DENSITY ---
        \multirow{4}{*}{\textbf{Density}} & \multirow{2}{*}{Buoyancy} & 
        \textbf{Relative Depth:} Object center relative to liquid height ($h_{\text{liq}} = y_{\text{liq}}^{\max} - y_{\text{liq}}^{\min}$). & 
        Float: $< 0.40$ & 
        \multirow{2}{*}{MLLM Text} \\
         & & \textit{Ratio:} $(y_{\text{obj}}^{\text{center}} - y_{\text{liq}}^{\min}) / h_{\text{liq}}$ & Sink: $> 0.75$ & \\
        \cmidrule(l){2-5}
         & \multirow{2}{*}{Balance} & 
        \textbf{Vertical Tilt:} Difference between left ($y_L^{\max}$) and right ($y_R^{\max}$) pan bottoms. & 
        \multirow{2}{*}{$\epsilon_{\text{bal}} = 20\text{px}$} & 
        \multirow{2}{*}{MLLM Text} \\
         & & \textit{Diff:} $|\, y_L^{\max} - y_R^{\max} \,| > \epsilon_{\text{bal}} \rightarrow \text{Tilt}$ & & \\
        \midrule
        
        % --- OOD ---
        \multirow{4}{*}{\textbf{OOD}} & \multirow{2}{*}{Size Reversal} & 
        \textbf{Area Ratio:} Area of giant object ($A_{\text{giant}}$) vs. tiny object ($A_{\text{tiny}}$). & 
        \multirow{2}{*}{Ratio $> 1.2$} & 
        \multirow{2}{*}{MLLM Text} \\
         & & \textit{Ratio:} $A_{\text{giant}} / A_{\text{tiny}}$ & & \\
        \cmidrule(l){2-5}
         & \multirow{2}{*}{Containment} & 
        \textbf{Intersection over Area (IoA):} Overlap area relative to inner object area ($A_{\text{inner}}$). & 
        \multirow{2}{*}{IoA $> 0.5$} & 
        \multirow{2}{*}{MLLM Text} \\
         & & \textit{Formula:} $A_{\text{overlap}} / A_{\text{inner}}$ & & \\
        \bottomrule
    \end{tabular}
\end{table*}

\begin{figure*}[t]
    \centering
    \includegraphics[width=0.95\linewidth]{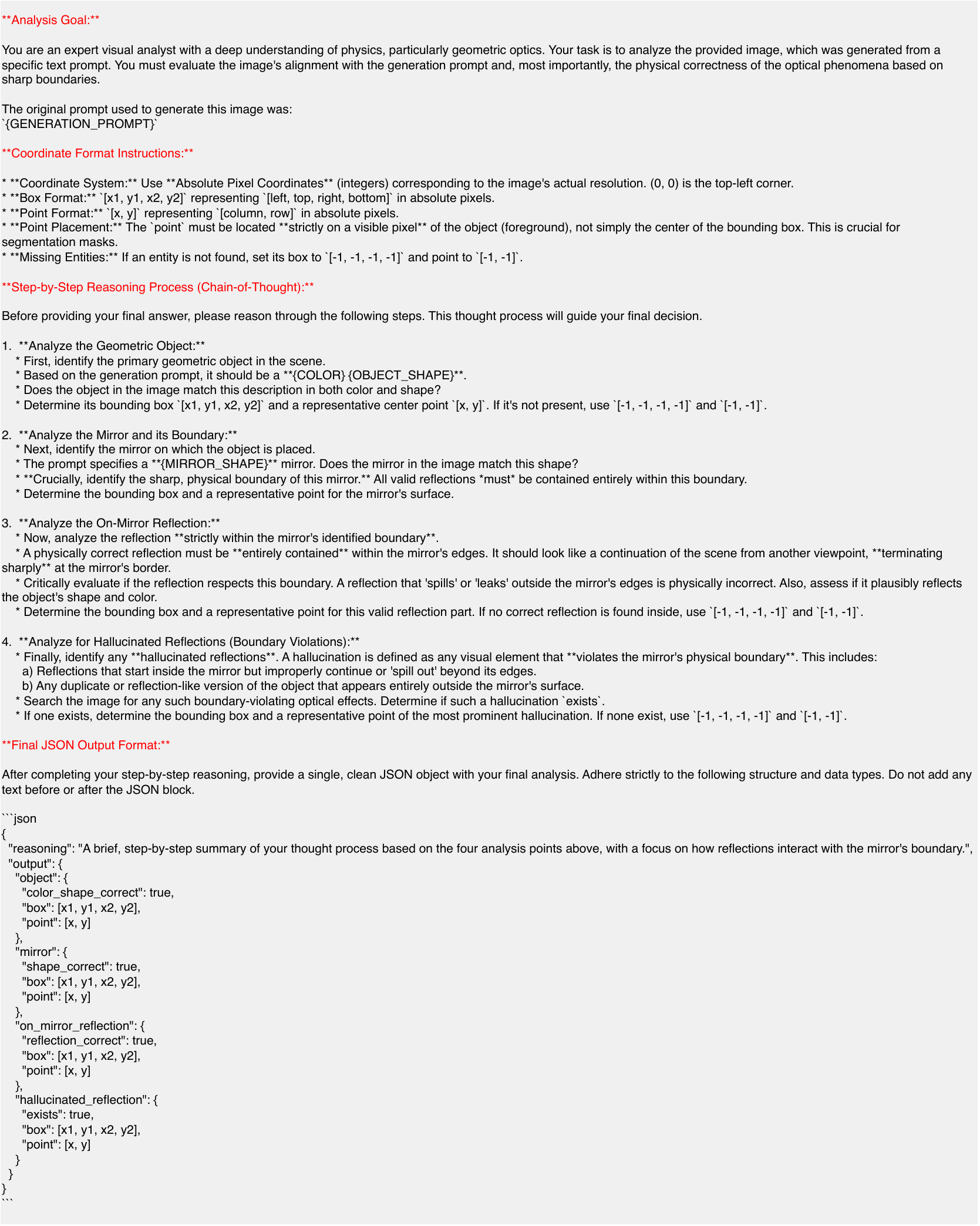} 
    \caption{\textbf{The System Prompt for the VLM-based Visual Parser (Optics Domain).} 
    This figure illustrates the full instruction set provided to Qwen2.5-VL-72B-Instruct. It includes strict coordinate formatting rules, a step-by-step chain-of-thought reasoning requirement, and the final JSON output schema tailored for analyzing reflection consistency.}
    \label{fig:vlm_prompt_optics}
\end{figure*}

To ensure a rigorous and reproducible assessment, we establish a standardized evaluation protocol that decouples visual perception from logical verification. We employ \textbf{Qwen2.5-VL-72B}~\citep{bai2025qwen2} as a visual parser to extract structured attributes (\ie, bounding boxes and object states) into JSON format. The specific system instruction and chain-of-thought prompting strategy used to guide the VLM in the Optics domain are illustrated in Figure~\ref{fig:vlm_prompt_optics}. To mitigate the stochasticity inherent in MLLM reasoning, we do not rely solely on the model's subjective text output. Instead, we implement a hybrid verification pipeline that prioritizes deterministic, rule-based geometric checks based on the extracted bounding boxes, falling back to semantic verification only when geometric constraints are inconclusive. We summarize the specific geometric formulas, thresholds ($\epsilon$), and logic flows for each domain in Table~\ref{tab:eval_criteria}.

\subsection{Evaluation on Optics}
In the Optics domain, accurate reflection generation requires both semantic correctness and geometric validity. We adopt the following criteria:
\begin{itemize}[leftmargin=*]
    \item \textbf{Semantic Consistency:} The visual parser must explicitly identify the presence of a reflection on the mirror surface and confirm its visual consistency with the main object.
    \item \textbf{Geometric Constraint:} We quantify the ``Boundary Violation'' error to filter out physically impossible reflections. Let $y_{\text{mirror}}^{\max}$ and $y_{\text{refl}}^{\max}$ denote the bottom-edge coordinates of the mirror and the reflection bounding boxes, respectively. A generation is penalized if the reflection violates the containment constraint:
    \begin{equation}
        y_{\text{refl}}^{\max} > y_{\text{mirror}}^{\max} + \epsilon_{\text{spill}}
    \end{equation}
    where $\epsilon_{\text{spill}}$ is a tolerance margin set to 15 pixels to accommodate minor segmentation noise. A sample is deemed successful only if it passes both checks.
\end{itemize}

\subsection{Evaluation on Density}
Evaluating physical interactions such as buoyancy and balance requires reference to ground-truth physical laws. We define the ``expected state'' based on the material densities specified in the dataset metadata. The evaluation follows a hierarchical logic:
\begin{itemize}[leftmargin=*]
    \item \textbf{Geometric Verification:} We primarily determine the object state using heuristic rules derived from bounding boxes. For \textit{Buoyancy}, we calculate the relative position of the object's center within the liquid container (normalized depth thresholds: top 40\% for floating vs. bottom 25\% for sinking). For \textit{Balance}, we compare the bottom-edge $y$-coordinates of the pans; a tilt is detected if the vertical difference exceeds $\epsilon_{\text{bal}} = 20$ pixels.
    \item \textbf{Semantic Fallback:} If the geometric configuration is ambiguous (\eg, due to obscured views), we fall back to the textual description provided by the MLLM. The final determined state is compared against the ground-truth metadata.
\end{itemize}

\subsection{Evaluation on Out-of-Distribution Scenarios}
For Out-of-Distribution (OOD) scenarios, success is defined by the model's adherence to explicit instruction constraints that contradict statistical priors. We employ specific verification strategies for different OOD types:
\begin{itemize}[leftmargin=*]
    \item \textbf{Spatial Constraints:} For \textit{Size Reversal} and \textit{Impossible Containment}, we rely on geometric metrics. Specifically, we verify that the area ratio satisfies $A_{\text{giant}} / A_{\text{tiny}} > 1.2$ for size reversal, and require the Intersection over Area (IoA) of the inner object relative to the container to exceed 0.5 for containment.
    \item \textbf{Semantic Constraints:} For \textit{Physics Violation} (\eg, ``an iron block floating''), where geometric priors might be misleading, we verify performance by comparing the MLLM-parsed observation against the ``forced state'' specified in the prompt metadata. This confirms that the DM has successfully prioritized the user instruction over its internal physical knowledge.
\end{itemize}

\section{Automated Segmentation Mask Generation}
\label{sec:appendix_mask_generation}

To facilitate the fine-grained diagnostic interventions described in Sec.~\ref{sec:diagnostics}, we generate high-precision binary masks for all semantic elements identified by the MLLM. We employ the \textbf{Segment Anything Model 2 (SAM 2)} architecture \citep{ravi2024sam}, utilizing the Hiera-Large backbone for optimal segmentation quality.

The generation pipeline automates the prompt engineering for SAM 2 by parsing the structured JSON outputs from our MLLM annotation step:
\begin{itemize}
    \item \textbf{Coordinate Rescaling and Prompting:} The MLLM perceives images at a standardized resolution ($512 \times 512$), whereas the generated images are typically high-resolution (\eg, $1024 \times 1024$). To bridge this domain gap, we extract the bounding box (\texttt{box}) and representative point (\texttt{point}) coordinates for each valid entity (\eg, ``\texttt{object}'', ``\texttt{mirror}'') and rescale them by the resolution ratio (\eg, $\times 2.0$). These rescaled spatial prompts are fed simultaneously into the SAM 2 predictor to maximize alignment.
    
    \item \textbf{Conditional Logic:} We implement strict validity checks to minimize noise. Masks are generated solely for entities with valid coordinates. Furthermore, specific causal elements are processed conditionally; for instance, a mask for ``\texttt{hallucinated\_reflection}'' is generated if and only if the MLLM explicitly flags its existence (\texttt{exists: true}).
    
    \item \textbf{Automated Mask Selection:} SAM 2 inherently addresses segmentation ambiguity by generating multiple candidate masks for a single prompt. To automate selection, we utilize the model's \textbf{predicted IoU score}, which serves as an internal confidence metric. We strictly select the candidate mask with the highest predicted score to ensure the segmentation best adheres to the provided spatial prompts.
\end{itemize}

This process results in a comprehensive dataset of high-quality segmentation masks, enabling the precise element-level manipulation required for our interventional experiments.

\section{Evaluation on Video Generation}
\label{sec:appendix_video_eval}

\subsection{Implementation Details}
We integrate \LINA with Wan-2.2-T2V-A14B \citep{wan2025wan}. The generation configuration is set to a resolution of $832 \times 480$ with a frame rate of 16 fps, yielding a total of 81 frames.
For the automated evaluation pipeline, we employ Qwen2.5-VL-72B \citep{bai2025qwen2}. To manage context length and computational efficiency, we downsample the generated videos to 4 fps before feeding them into the MLLM. This sampling rate is sufficient to capture key temporal dynamics and physical state changes without redundancy.

\subsection{Domain Gap Analysis}
Extending causal interventions from image to video reveals specific domain shifts beyond the temporal dynamics discussed in the main text.

\paragraph{Sensitivity to Verbs vs. Nouns.}
We observe that video DMs exhibit a heightened sensitivity to verbs (actions) compared to image DMs, which are predominantly sensitive to nouns (objects). Consequently, for the PAP-Density adaptation, we prioritized prompts involving physical state changes (\eg, sinking, floating) over static spatial arrangements. This ensures the evaluation isolates physical reasoning from camera movement or narrative hallucinations.

\paragraph{Geometric Abstraction vs. Realism.}
A notable limitation of the Wan-2.2 model is its bias towards photorealistic textures over abstract geometric primitives. For instance, when prompted with ``a red sphere'', the model frequently generates semantically related real-world objects such as tomatoes or strawberries. This domain shift results in lower baseline scores on geometry-heavy tasks compared to image DMs, as the model struggles to decouple the geometric shape from its learned semantic texture associations.

{
    \small
    \bibliographystyle{ieeenat_fullname}
    \bibliography{main}
}

\end{document}